\documentclass[10pt,twocolumn,letterpaper]{article}

\usepackage[pagenumbers]{cvpr} 
\usepackage{bm}
\usepackage{booktabs}
\usepackage{bm}
\usepackage{tabularx}
\usepackage{enumitem,kantlipsum}
\usepackage{multirow}
\usepackage[usestackEOL]{stackengine}
\usepackage[ruled,vlined]{algorithm2e}
\usepackage{graphicx}
\usepackage[most]{tcolorbox}
\usepackage{tikz}

\usepackage{wrapfig}  

\definecolor{skyblue}{RGB}{232, 243, 255}

\newcommand{\cryptoface}{CryptoFace}
\newcommand{\cryptofacenet}{CryptoFaceNet}

\newcommand{\ourface}{CryptoFace}

\usepackage{bm}
\usepackage{tabularx}
\usepackage{enumitem,kantlipsum}
\usepackage{multirow}
\usepackage[usestackEOL]{stackengine}
\usepackage[ruled,vlined]{algorithm2e}
\usepackage{graphicx}
\usepackage[most]{tcolorbox}
\usepackage{color, colortbl}
\usepackage{listings}

\usepackage{amsmath,amsfonts,bm}

\def\Figref#1{Figure~\ref{#1}}

\def\eqref#1{equation~\ref{#1}}
\def\Eqref#1{Equation~\ref{#1}}

\def\1{\bm{1}}

\DeclareMathAlphabet{\mathsfit}{\encodingdefault}{\sfdefault}{m}{sl}
\SetMathAlphabet{\mathsfit}{bold}{\encodingdefault}{\sfdefault}{bx}{n}

\RequirePackage{xspace}
\makeatletter
\DeclareRobustCommand\onedot{\futurelet\@let@token\@onedot}
\def\@onedot{\ifx\@let@token.\else.\null\fi\xspace}
\def\eg{\emph{e.g}\onedot} 

\def\ie{\emph{i.e}\onedot}

\definecolor{rulecolor}{RGB}{0,71,171}
\definecolor{tableheadcolor}{RGB}{204,229,255}

\newcommand{\topline}{ %
    \arrayrulecolor{rulecolor}\specialrule{0.1em}{\abovetopsep}{0pt}%
    \arrayrulecolor{rulecolor}\specialrule{\lightrulewidth}{0pt}{0pt}%
    \arrayrulecolor{tableheadcolor}\specialrule{\aboverulesep}{0pt}{0pt}%
    \arrayrulecolor{rulecolor}
    }
\newcommand{\midtopline}{
    \arrayrulecolor{tableheadcolor}\specialrule{\aboverulesep}{0pt}{0pt}%
    \arrayrulecolor{rulecolor}\specialrule{\lightrulewidth}{0pt}{0pt}%
    \arrayrulecolor{white}\specialrule{\aboverulesep}{0pt}{0pt}%
    \arrayrulecolor{rulecolor}}
    \newcommand{\bottomline}{
    \arrayrulecolor{tableheadcolor}\specialrule{\aboverulesep}{0pt}{0pt}
    \arrayrulecolor{rulecolor}
    \specialrule{\heavyrulewidth}{0pt}{\belowbottomsep}
    \arrayrulecolor{rulecolor}\specialrule{\lightrulewidth}{0pt}{0pt}
    }
\newcolumntype{?}{!{\vrule width 1.4pt}}

\definecolor{cvprblue}{rgb}{0.21,0.49,0.74}
\usepackage[pagebackref,breaklinks,colorlinks,allcolors=cvprblue]{hyperref}

\def\confName{CVPR}
\def\confYear{2025}

\title{\ourface{}: End-to-End Encrypted Face Recognition}

\author{Wei Ao \quad Vishnu Naresh Boddeti\\
Michigan State University\\
{\tt\small \{aowei, vishnu\}@msu.edu} \\
}

\begin{document}

\maketitle
\begin{abstract}
    Face recognition is central to many authentication, security, and personalized applications. Yet, it suffers from significant privacy risks, particularly arising from unauthorized access to sensitive biometric data. This paper introduces CryptoFace, the first end-to-end encrypted face recognition system with fully homomorphic encryption (FHE). It enables secure processing of facial data across all stages of a face-recognition process—feature extraction, storage, and matching—without exposing raw images or features. We introduce a mixture of shallow patch convolutional networks to support higher-dimensional tensors via patch-based processing while reducing the multiplicative depth and, thus, inference latency. Parallel FHE evaluation of these networks ensures near-resolution-independent latency. On standard face recognition benchmarks, CryptoFace significantly accelerates inference and increases verification accuracy compared to the state-of-the-art FHE neural networks adapted for face recognition. CryptoFace will facilitate secure face recognition systems requiring robust and provable security. The code is available at \url{https://github.com/human-analysis/CryptoFace}. 
\end{abstract}
\section{Introduction}

\begin{figure}
    \centering
    \includegraphics[width=\linewidth]{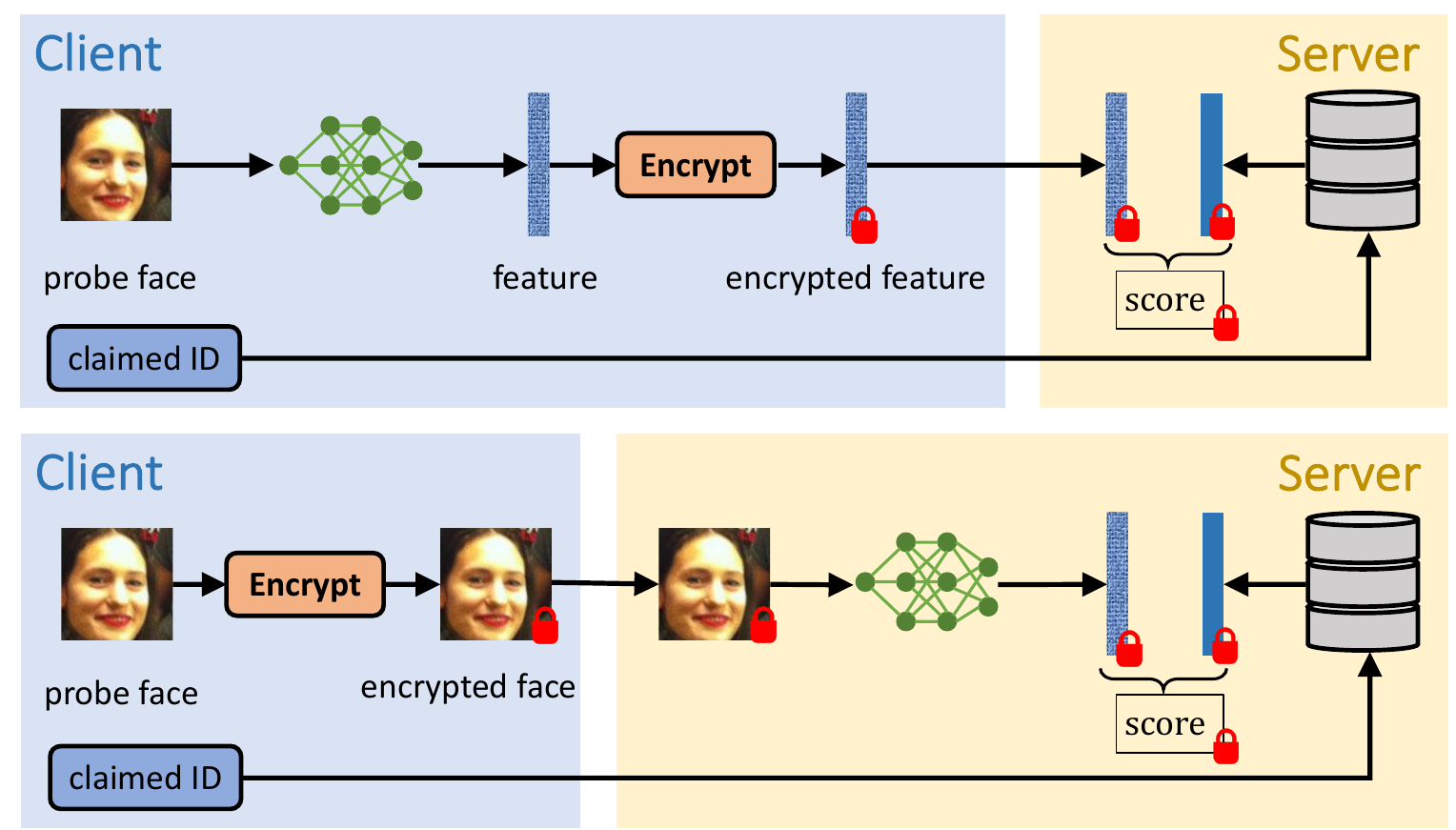}
    \caption{Secure FR systems. (1) Top: existing secure FR system which encrypts only the features, rather than raw face images, limits practical utility and security; (2) bottom: proposed end-to-end encrypted face recognition system which ensures stronger protection of the user's face image and feature while safeguarding the server’s reference database.\label{fig:enc_fr}} 
    \vspace{-0.5cm}
\end{figure}

\textbf{Face Recognition (FR)}~\cite{wang2018cosface,deng2022arcface,kim2022adaface} has become integral to identity management in many practical applications, from unlocking personal devices to facilitating law enforcement and accessing financial services. These systems process sensitive biometric data that, if compromised, can lead to privacy invasions, identity theft, and unauthorized surveillance. Unlike passwords, biometric data is immutable—once compromised, it cannot be changed, which elevates the need for robust security mechanisms to protect it. Such protections are also mandated by legal regulations on the acquisition, storage, and usage of biometric data~\cite{meden2021privacy}, \eg the European Union's General Data Protection Regulation (GDPR)~\cite{voigt2017eu}. FR systems in the wild consist of three entities: a probe face image, a feature extractor (\ie a FR neural network), and a reference database of face features. The feature extractor processes a face image to generate its corresponding compact feature representation. 

\noindent \textbf{Secure FR Systems.} FR systems meet increasing security challenges. An adversarial client could attempt to infer biometric data from the server's reference database or leverage the feature extractor to generate adversarial probes to deceive the system. Similarly, an adversarial server might exploit client-provided data to extract biometric information or infer sensitive attributes such as gender, age, or race. Existing secure FR systems apply homomorphic encryption (HE) in a client-server two-party scenario~\cite{boddeti2018secure,engelsma2022hers} to ensure security of \emph{face features}. HE allows computation over encrypted data and provides provable post-quantum security~\cite{gentry2009ideal, gentry2009fhe,lyubashevsky2010ideal,cheon2017homomorphic,
cheon2018full}. As shown in \Cref{fig:enc_fr} (top), the client's probe feature and the reference database at a server are encrypted. The client performs feature extraction locally and cannot delegate this task to the server. Verification is performed by the server directly within the encrypted domain. Limiting security measures to \emph{feature protection only} reduces the practical utility of such systems. Moreover, such a secure FR system is vulnerable to a template recovery attack~\cite{bassit2023template} where an adversarial client could attempt to infer features in the server's reference database. Unlike HE, which provides robust and provable security guarantees, other lines of research in privacy-preserving face recognition focus on different security or privacy issues, such as adversarial attacks~\cite{deb2020advfaces}, viewing attacks~\cite{mi2024privacy}, and information leakage~\cite{ji2022privacy}.

\noindent \textbf{\ourface{}.} To address security vulnerabilities in existing secure FR systems, we introduce \ourface{}, the \emph{first end-to-end encrypted} FR system using fully homomorphic encryption (FHE). All operations--including feature extraction, feature matching, and score comparison--are performed entirely within the encrypted domain without decryption at any point. As shown in \Cref{fig:enc_fr} (bottom), the client encrypts the probe face image and sends it to the server. The server uses a neural network (NN) to extract an encrypted feature. This encrypted feature is matched against the encrypted reference feature stored in the server's database, producing an encrypted similarity score. The score is compared against a threshold, and the encrypted match/no-match result is returned to the client, who alone can decrypt it. By securing the user's sensitive biometric data throughout the verification process and protecting the server's reference database, \ourface{} offers robust end-to-end security for FR systems. 

Realizing this goal presents three major technical challenges: (1) Homomorphic evaluation of state-of-the-art (SoTA)  convolutional neural networks (CNNs) is computationally demanding due to the high multiplicative depth of CNNs~\cite{cheon2017homomorphic,cheon2018full,lee2022low,ao2023autofhe}. (2) Although several approaches~\cite{gilad2016cryptonets,lee2022low,ao2023autofhe} have demonstrated homomorphic evaluation of CNNs on low-resolution images, they cannot directly process higher-resolution face images. (3) FR requires the cosine similarity measure~\cite{deng2022arcface}, which cannot be directly computed on FHE. Existing secure FR~\cite{boddeti2018secure,engelsma2022hers} circumvented the evaluation of the cosine function under FHE by normalizing the features before encryption.

\noindent\textbf{\cryptofacenet{}} is designed to address these technical challenges. A face image is divided into a grid of non-overlapping patches~\cite{dosovitskiy2020image}, and each is processed independently by a shallow patch CNN (PCNN). The mixture of PCNNs is jointly trained to learn the inter-patch relationships. Due to the lower resolution of individual patch, we reduce the multiplicative depth required for each PCNN. We also optimize convolutional blocks to further minimize the multiplicative depth and adopt other recent advancements for efficient FHE convolution~\cite{lee2022low} and low-degree polynomial~\cite{park2022aespa} activation functions. Under FHE, the mixture of PCNNs is evaluated in parallel and features are additively aggregated, significantly accelerating the feature extraction process. \cryptofacenet{} scales effectively to high-resolution face images and maintains near-resolution-independent latency due to parallelism. Additionally, we design a distribution-aware low-degree polynomial approximation of the cosine similarity function to efficiently compute the similarity score under FHE. 

We evaluate \cryptoface{} on standard FR benchmarks, comparing its performance to SoTA FHE CNNs~\cite{lee2022low, ao2023autofhe} adapted for FR. Our results show that \cryptoface{} not only improves the one-to-one verification accuracy by up to $+8.8\%$ but also speeds up the encrypted FR by $7\times$. \cryptoface{} supports arbitrary-resolution images, maintaining near-constant latency across different resolutions. We summarize the \textbf{contributions} of this paper below:
\begin{tcolorbox}[width=\linewidth,colback=skyblue,colframe=skyblue,boxsep=0pt,left=3pt,right=3pt]
\begin{enumerate}[leftmargin=*]
    \item \textbf{End-to-End Encrypted Face Recognition:} \cryptoface{} is the first secure FR system to perform feature extraction, feature matching, and score thresholding entirely within the encrypted domain, eliminating the need for decryption at any stage. \cryptoface{} enhances security and expands the applicability of secure FR.
    \item \textbf{Efficient and Scalable Architecture:} \cryptofacenet{} is a novel FHE-compatible architecture that reduces computational overhead by minimizing multiplicative depth and is scalable to high-resolution face images. 
    \item \textbf{Feasibility and Efficacy Demonstration:} We present the first practical implementation of end-to-end encrypted face recognition, demonstrating its feasibility on standard FR benchmarks under FHE.
\end{enumerate}
\end{tcolorbox} 

\section{Background and Related Work \label{sec:related}}

\subsection{Homomorphic Encryption (HE) \label{sec:HE}}

Homomorphic encryption (HE) is a class of
encryption schemes that are considered quantum-secure and enable computations on
encrypted data without requiring decryption. HE schemes are based on the Learning
with Errors (LWE) problem~\cite{gentry2009ideal, gentry2009fhe} or Ring Learning with Errors (RLWE)~\cite{lyubashevsky2010ideal}. Among different HE schemes, the Cheon-Kim-Kim-Song (CKKS) encryption scheme~\cite{cheon2017homomorphic,cheon2018full,cheon2018bootstrapping} is particularly well-suited for encrypted inference in neural networks
since it supports fixed-point approximate arithmetic over complex and real numbers. 

\noindent\textbf{Encryption and Decryption.} A \emph{cleartext} message vector $\mu \in \mathbb{C}^{\frac{N}{2}}$ is first encoded into a \emph{plaintext} message $m$, which is subsequently
encrypted into a \emph{ciphertext} $\bm{c}$ using a public key
$pk=(-\langle a,sk \rangle+e, a)$. Here, $\langle \cdot, \cdot \rangle$ is dot product operator, $a$ is a random ring, $e$ is an encryption noise, and $sk$ is
the secret key. The encryption and decryption processes are defined as follows:
\begin{align}
  \begin{split}&\mathrm{Encrypt}(m, pk) = (m, 0) + pk = (c_{0},c_{1}) = \bm{c}\\&\mathrm{Decrypt}(\bm{c}, sk) = c_{0} + \langle c_{1}, sk \rangle=m+e\end{split} \label{eq:ckks}
\end{align}
The CKKS scheme uses a residue cyclotomic polynomial ring $\mathcal{R}_{Q_\ell}=\mathbb{Z}
_{Q_\ell}[X]/(X^{N}+1)$ to encode cleartext vectors. The modulus is defined as
$Q_{\ell}=\prod_{i=0}^{\ell} q_{\ell}, 0 \leq \ell \leq L$. The polynomial degree
$N$ determines the message capacity, allowing $\frac{N}{2}$ complex numbers to
be packed into $\frac{N}{2}$ slots.

\noindent\textbf{Supported Operations.} The CKKS scheme supports two homomorphic operations—addition and multiplication—and one automorphic operation, rotation. These operations are defined as follows~\cite{cheon2017homomorphic,lee2022low}:
\begin{align}
  \begin{split}\underbrace{[m_1]\oplus [m_2]}_{\text{ciphertext-ciphertext}}&\approx \underbrace{[m_1] \oplus m_2}_{\text{ciphertext-plaintext}}\approx [\underbrace{m_1 + m_2}_{\text{element-wise add}}] \\ \underbrace{[m_1] \otimes [m_2]}_{\text{ciphertext-ciphertext}}&\approx \underbrace{[m_1] \otimes m_2}_{\text{ciphertext-plaintext}}\approx [\underbrace{m_1 \times m_2}_{\text{element-wise mul}}] \\ \mathrm{Rot}([m], r)&= [\mathrm{Rot}(m, r)] \\\end{split}
  \label{eq:operation}
\end{align}
Here, $[\cdot]$ represents an encrypted message or vector and $\mathrm{Rot}(\cdot)$ denotes a left cyclical rotation of the vector by $r$ positions. Homomorphic addition ($\oplus$) and multiplication ($\otimes$) can be applied between two ciphertexts or between a ciphertext and plaintext, enabling element-wise computations.


\noindent\textbf{Multiplicative Level and Depth.} Each ciphertext is associated with a \colorbox{skyblue}{level} $\ell$, an integer indicating the number of homomorphic multiplications that can be performed before decryption fails. A function with a multiplicative \colorbox{skyblue}{depth} $k$--defined as the number of sequential homomorphic multiplications it involves—consumes $k$ levels. After each multiplication, the polynomial modulus $Q_\ell$  must be rescaled, transitioning from  $Q_\ell$ to $Q_{\ell - 1}$ to maintain the scale~\cite{cheon2017homomorphic}.

\noindent\textbf{LHE and FHE.} CKKS without bootstrapping is a leveled homomorphic encryption (\colorbox{skyblue}{LHE}) scheme, allowing a limited number of multiplications determined by the initial level $L$ of a freshly encrypted ciphertext. To evaluate functions with arbitrary depth, fully homomorphic encryption (\colorbox{skyblue}{FHE}) incorporates a \colorbox{skyblue}{bootstrapping} operation~\cite{cheon2018bootstrapping, bossuat2021efficient, lee2021high} to \emph{refresh} ciphertexts, effectively resetting their level to enable further computation. For deeper neural networks, bootstrapping must be periodically applied to prevent decryption failures. However, this process is computationally expensive, with high latency due to the large number of rotation operations involved. Additionally, bootstrapping has a significant memory footprint because of the large size of the bootstrapping operators. While a freshly encrypted ciphertext starts with $L$ levels, bootstrapping reduces the available levels to $L - K$, as $K$ levels are consumed during the evaluation of polynomials required for bootstrapping. 

\noindent\textbf{Computational Complexity.} Bootstrapping is slower than rotation or ciphertext-ciphertext multiplication by two orders of magnitude~\cite{kim2023hyphen}. Ciphertext-ciphertext multiplication is slower than ciphertext-plaintext multiplication or addition by two orders of magnitude~\cite{kim2023hyphen}. 

\subsection{Homomorphic CNNs (HCNNs) \label{sec:hcnns}}
Homomorphic CNNs (HCNNs) are CNNs that are compatible with the operations that HE supports in~\Eqref{eq:operation}. We categorize HCNNs into \colorbox{skyblue}{FHENets} and \colorbox{skyblue}{LHENets} depending on whether bootstrapping is used
or not, respectively. LHENets include CryptoNets~\cite{gilad2016cryptonets},
LoLa~\cite{brutzkus2019low}, Faster CryptoNets~\cite{chou2018faster},
while recent FHENets include MPCNN~\cite{lee2022low} and
AutoFHE~\cite{ao2023autofhe}. FHENets achieve SoTA prediction accuracy on
image classification datasets but with much higher latency than LHENets. Existing
HCNNs to speed up the inference on encrypted images have focused on two aspects, packing for convolutions and polynomial activation.

\noindent \textbf{Packing for Convolutions} refers to efficiently packing three-dimensional tensors to reduce the complexity of HE multiplication and rotation for convolutional layers~\cite{juvekar2018gazelle,lee2022low}. MPCNN designs multiplexed convolution by integrating (1) \colorbox{skyblue}{repeated packing} and (2) \colorbox{skyblue}{multiplexed packing}~\cite{lee2022low}. (1) A ciphertext with the cyclotomic polynomial degree $N$ can pack $\frac{N}{2}$ numbers. Given a vector $x\in\mathbb{R}^d$ with $d<\frac{N}{2}$, a repeated vector is $x^{(M)} = [x, x, \cdots, x]$ with $M=\lfloor \frac{N}{2d} \rfloor$ copies of $x$. $x^{(M)}$ is encrypted to fill out all slots. The repeated packing can accelerate convolution and bootstrapping operations. A larger $M$ leads to faster inference. (2) When the convolutional stride exceeds 1, gaps between valid values are introduced, causing some ciphertext slots to remain unused. MPCNN addresses this issue by packing numbers from different channels into alternate slots, effectively filling these gaps and fully utilizing all ciphertext slots, thereby preventing sparsity. Additionally, channels are computed in parallel to speed up convolutional layers.

\noindent \textbf{Polynomial Activation.} HCNNs cannot employ $\mathrm{ReLU}$ since it is not a homomorphism. So, they adopt polynomial activations (e.g., monomial, Chebyshev or Hermite polynomials) to replace $\mathrm{ReLU}$. Monomial polynomials are widely used since they allow HCNNs to be formulated as traditional polynomial networks. Examples include CryptoNets~\cite{gilad2016cryptonets}, LoLa~\cite{brutzkus2019low}, and Faster CryptoNets~\cite{chou2018faster}. However, training with monomial polynomials often becomes unstable due to exploding gradients. Minimax approximations using Chebyshev polynomials can achieve high-precision approximations of ReLU functions~\cite{lee2021minimax,lee2021precise,lee2022low}, but their high polynomial degree leads to prohibitively large multiplicative depths. AESPA~\cite{park2022aespa} addresses this issue by introducing low-degree Hermite polynomials to reduce the multiplicative depth and proposes basis-wise normalization to stabilize training. Furthermore, search-based AutoFHE~\cite{ao2023autofhe} explores layer-wise mixed-degree polynomials to further decrease multiplicative depth.

\noindent \textbf{\cryptofacenet{} Modules.} We build upon the above mentioned prior advances for accelerating HCNN inference, specifically adopting multiplexed convolution~\cite{lee2022low} and low-degree, basis-wise normalized Hermite activation~\cite{park2022aespa}. Our primary focus, however, lies in addressing other outstanding challenges of end-to-end encrypted FR, including processing high-resolution encrypted images and minimizing the multiplicative depth to reduce computationally expensive bootstrapping operations. We overcome these challenges through \cryptofacenet{}, a novel architecture design.
\section{End-to-End Encrypted FR System \label{sec:approach}}

\subsection{\ourface{} \label{sec:cryptoface}}

\textbf{FR Models}~\cite{wang2018cosface,deng2022arcface,kim2022adaface} take advantage of the outstanding representation learning ability of CNNs~\cite{he2016deep} to extract discriminative features. Given a face image $x\in\mathbb{R}^{C\times H \times W}$, a neural network $f_\omega(\cdot)$ with trainable parameters $\omega$, we have feature $y=f_\omega(x)\in\mathbb{R}^d$. To verify if two face images $x_1$, $x_2$ belong to the same individual, we compare their features $y_1=f_\omega(x_1)$ and $y_2=f_\omega(x_2)$ to obtain their similarity, \ie $\operatorname{Score}(y_1, y_2) = \| \frac{y_1}{\|y_1 \|} - \frac{y_1}{\|y_1 \|} \|^2 = 2 - 2\frac{y_1 y_2}{\|y_1\| \| y_2 \|}$. If $\operatorname{Score}(y_1, y_2)$ is smaller than a predefined threshold $\operatorname{T}$, the two face images are classified as corresponding to the same person. We formulate this process as a $\mathrm{Match}$ function:
\begin{equation}
    \mathrm{Match}(y_1,y_2) = \operatorname{Score}(y_1, y_2) - \mathrm{T}
    \label{eq:match}
\end{equation}
A neural network $f_\omega(\cdot)$ for FR is trained on a dataset $\mathcal{X}$ with $M$ identities, each consisting of multiple face images. To train, we need to learn the feature center $W\in\mathbb{R}^{M\times d}$ with $M$ $d$-dimensional feature centers, where $d$ is the predefined feature dimension. We use ArcFace's~\cite{deng2022arcface} additive angular margin loss which is a modified cross-entropy loss:
\begin{equation}
    \resizebox{0.9\linewidth}{!}{$\mathcal{L}_{\mathrm{ArcFace}}(\omega)= - \log \frac{e^{s \cos{(\theta_{i}+m)}}}{e^{s \cos{(\theta_{i}+m)}}+\sum_{j=1,j\neq i}^M e^{s \cos{\theta_j}}}$}
    \label{eq:arcface}
\end{equation}
\noindent
Given a training sample $x$ with identity $i$, the feature $y=f_\omega(x)$. The cosine similarity values between $y$ and $M$ $d$-dimensional features of the feature center are obtained. In Equation~\ref{eq:arcface}, $\theta_i$ is the angle between $y$ and $M[i]$, while $\theta_j$ is the angle between $y$ and $M[j], j\neq i$. The additive margin $m$ is added to the angle $\theta_i$. The scalar $s$ increases the capacity of the unit ball. 

Face verification involves two phases, \emph{enrollment} and \emph{verification}. In the enrollment stage, the client sends the server a reference face image and corresponding identity. The server employs the trained network to extract the feature from the reference face image and stores the feature and identity. In the verification stage, the client sends a probe face image and a claimed identity to the server. The server employs the same network to extract the feature from the probe face image and compares it with the reference feature indexed by the claimed identity.

\begin{figure}[t]
    \centering
    \includegraphics[width=\linewidth]{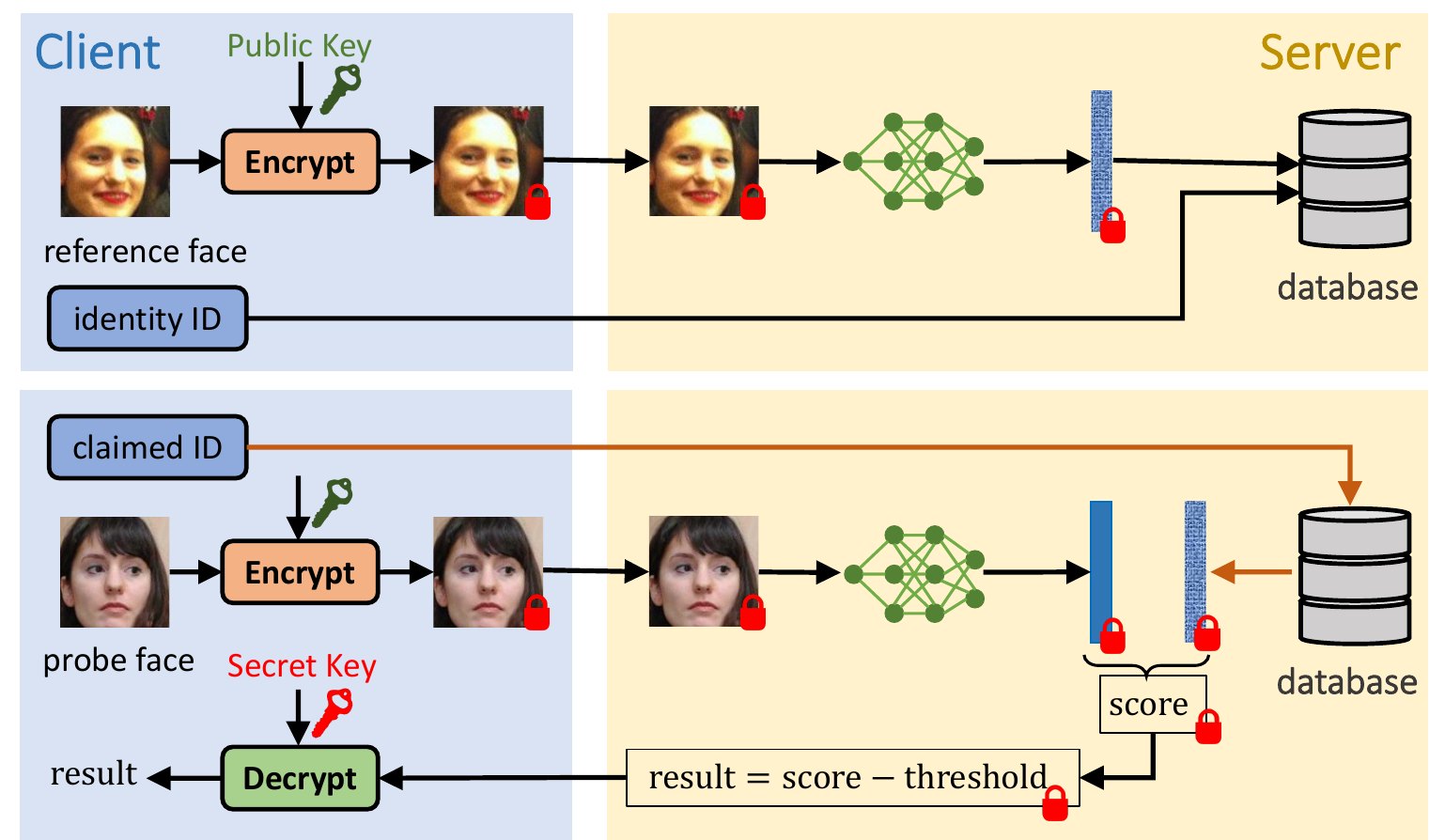}
    \caption{\ourface{}. Top: offline; bottom: online. \label{fig:end-to-end secure FR}}
    \vspace{-0.5cm}
\end{figure}

\noindent\textbf{Encrypted FR} comprises two similar phases in our paper, \emph{offline} and \emph{online} as shown in~\Cref{fig:end-to-end secure FR}. The offline stage is similar to enrollment, and the online stage is analogous to verification. In the offline stage, the client generates a public key to encrypt the reference face image and sends the encrypted reference face image and the corresponding identity to the server. The server extracts the encrypted feature $[y_1]=f_\omega([x_1])$. The client does not need to wait for the offline stage to complete. However, during the online stage, the client must wait for the inference result; thus, the latency of the online stage is critical for real-world applications. In the online stage, the client encrypts a probe image $[x_2]$ and sends the encrypted probe face image and a claimed identity to the server. The server extracts the encrypted feature $[y_2]=f_\omega([x_2])$. Then, the server computes the match function (\Eqref{eq:match}) over the encrypted features, i.e., $\mathrm{Match}([y_1], [y_2])$, and finally returns the resulting encrypted match result to the client. The client uses the secret key to decrypt the result and check if it is \colorbox{GreenYellow}{negative} or \colorbox{Lavender}{positive} to determine a \colorbox{GreenYellow}{match} or \colorbox{Lavender}{no match}, respectively. 

\vspace{3pt}\noindent\textbf{Threat Model.} Following the threat model used by existing encrypted FR systems~\cite{boddeti2018secure,engelsma2022hers} and recent FHENets~\cite{lee2022low,ao2023autofhe}, we assume two semi-honest parties: a client and a server. Under this model, at most one of these parties may be corrupted by an adversary~\cite{mishra2020delphi}. Although both parties follow the agreed-upon protocol honestly, they may attempt to extract additional information by analyzing the data received from each other~\cite{mishra2020delphi}.
\begin{itemize}[leftmargin=*]
    \item If \emph{the client is adversarial}, it may attempt to infer encrypted features stored on the server. However, since encrypted feature extraction is performed entirely on the server side without releasing intermediate features, the client receives only the matching result—a positive or negative scalar. Consequently, an adversarial client cannot infer the encrypted features stored on the server.
    \item If \emph{the server is adversarial}, it may attempt to collect biometric data from the client. However, since the server only holds encrypted face images and features without access to the client’s secret key, it cannot decrypt or infer any face images or features provided by the client. 
\end{itemize}

\subsection{\cryptofacenet{} \label{sec:mixture}}

Inspired by patch-based neural networks~\cite{dosovitskiy2020image}, \cryptofacenet{} applies a mixture of PCNNs to extract local features and fuse these local features to obtain a global feature. Such a design significantly reduces FHE latency due to shallow PCNNs with lower multiplicative depth and parallelized evaluation under FHE.

\begin{figure}
    \centering
    \includegraphics[width=\linewidth]{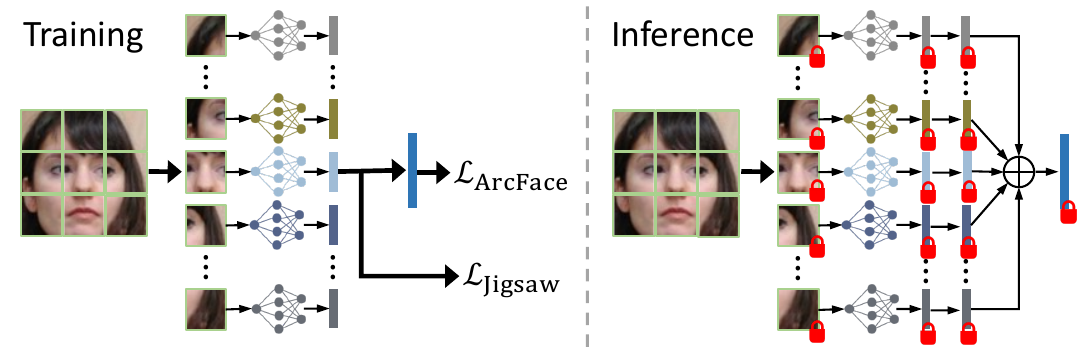}
    \caption{Left: cleartext training; right: encrypted inference. \label{fig:jigsaw_train}} 
    \vspace{-0.5cm}
\end{figure}

\noindent\textbf{Training} process on cleartext data is shown in~\Figref{fig:jigsaw_train} (left). A two-dimensional face image $x\in \mathbb{R}^{C\times H \times W}$ is divided into a sequence of patches ${x_i \in \mathbb{R}^{C\times P \times P}}{i=1}^{L}$, where $L=HW/P^2$. The image splitting approach follows that of Vision Transformer (ViT)~\cite{dosovitskiy2020image}. A mixture of PCNNs, denoted as $\{f_{\omega_i}\}_{i=1}^L$, independently processes each patch $x_i$ to generate local features $y_i\in\mathbb{R}^d$ for $1 \leq i \leq L$. These $L$ local features are subsequently fused to form the global feature utilized for FR as follows,
\begin{equation}
    y = y^\prime A^T + b,~\text{where}~y^\prime = [y_1, y_2, \cdots, y_L]
    \label{eq:fusion}
\end{equation}
where $A\in \mathbb{R}^{d\times dL}$ and $b\in \mathbb{R}^d$. The patch size is significantly smaller than the original image dimensions, i.e., $P\ll H$ or $P\ll W$, resulting in a reduced receptive field. This allows us to employ shallower PCNNs with lower multiplicative depth for each patch. Typically, each patch covers a small, distinctive facial structure, and the mixture of PCNNs learns separate filters for each patch instead of using weight-sharing~\cite{noroozi2016jigsaw}. The feature fusion process (\Eqref{eq:fusion}) encourages the PCNN mixture to capture inter-patch relationships effectively.

To train the mixture of PCNNs, we apply the ArcFace loss defined in~\Eqref{eq:arcface}. Additionally, we introduce a jigsaw puzzle auxiliary task~\cite{noroozi2016jigsaw,chen2023jigsaw,ren2023jigasw} to supplement positional information. This auxiliary task, previously shown to effectively capture positional details~\cite{noroozi2016jigsaw,chen2023jigsaw,ren2023jigasw}, naturally complements our mixture of PCNNs. Specifically, we use the local features $y_1, y_2, \cdots, y_L$ to predict their original positions (1, 2, $\cdots$, $L$) via a fully connected layer. Thus, the learned local features inherently encode positional information. The training objective of the mixture of PCNNs is:
\begin{equation}
    \mathcal{L}(\omega, W, A, b) = \mathcal{L}_{\mathrm{ArcFace}}(\omega, W, A, b) + \alpha \mathcal{L}_{\mathrm{Jigsaw}}(\omega)
    \label{eq:train_loss}
\end{equation}
where $\omega=\{\omega_i\}_{i=1}^L$ are the parameters of all PCNNs, $W$ is the feature center, $A$ and $b$ are feature fusion parameters, and $\alpha$ is the strength of jigsaw loss.

\noindent \textbf{Inference.} As illustrated in~\Cref{fig:jigsaw_train} (right), the acceleration achieved by evaluating an encrypted face using the proposed mixture of PCNNs is two-fold. First, the multiplicative depth is significantly reduced, requiring only one bootstrapping operation per PCNN. Second, the evaluation of these $L$ PCNNs can be performed in parallel, simplifying engineering implementation. However, the fusion step involves a large vector-matrix product $y^\prime A^T$ as in~\Eqref{eq:fusion}, which requires numerous computationally expensive homomorphic rotations. The mapping matrix $A \in \mathbb{R}^{d \times dL}$ is rectangular, making it incompatible with efficient HE vector-matrix multiplication approaches designed specifically for square matrices~\cite{halevi2014helib}. To address this issue, we rewrite $A$ as $A = [A_1, A_2, \cdots, A_L]$, where each $A_i \in \mathbb{R}^{d\times d}$ for $1 \leq i \leq L$. Consequently, the vector-matrix multiplication can be decomposed as follows: 
\begin{equation}
    y = \sum_{i=1}^L{\left(y_i A_i^T + \frac{b}{L}\right)},~\text{where}~y_i\in\mathbb{R}^d, A_i\in\mathbb{R}^{d\times d}
\end{equation}
\noindent where the vector-matrix product $y_iA_i^T+b/L$ can be evaluated in parallel. Under FHE, the fusion function reduces to a simple \emph{addition}, the computationally least expensive operation in FHE. 

\noindent\textbf{Scalability.} Existing HCNNs described in~\Cref{sec:hcnns} do not scale effectively to high-resolution face images. Increasing the size of tensors requires enlarging the degree ($N$) of the residue cyclotomic polynomial ring (see \Cref{sec:HE}), leading to a substantial accumulation of latency. This occurs because ciphertext multiplication and rotation complexity under FHE scales as $\mathcal{O}(\ell^2 N\log{N})$~\cite{lou2021hemet}. In contrast, the proposed mixture of PCNNs is highly scalable to higher resolutions. By increasing the number of PCNNs, our approach achieves near-resolution-independent inference speed, due to the novel and efficient parallel evaluation strategy we introduce.

\begin{figure*}[htb]
    \centering
    \includegraphics[width=0.9\linewidth]{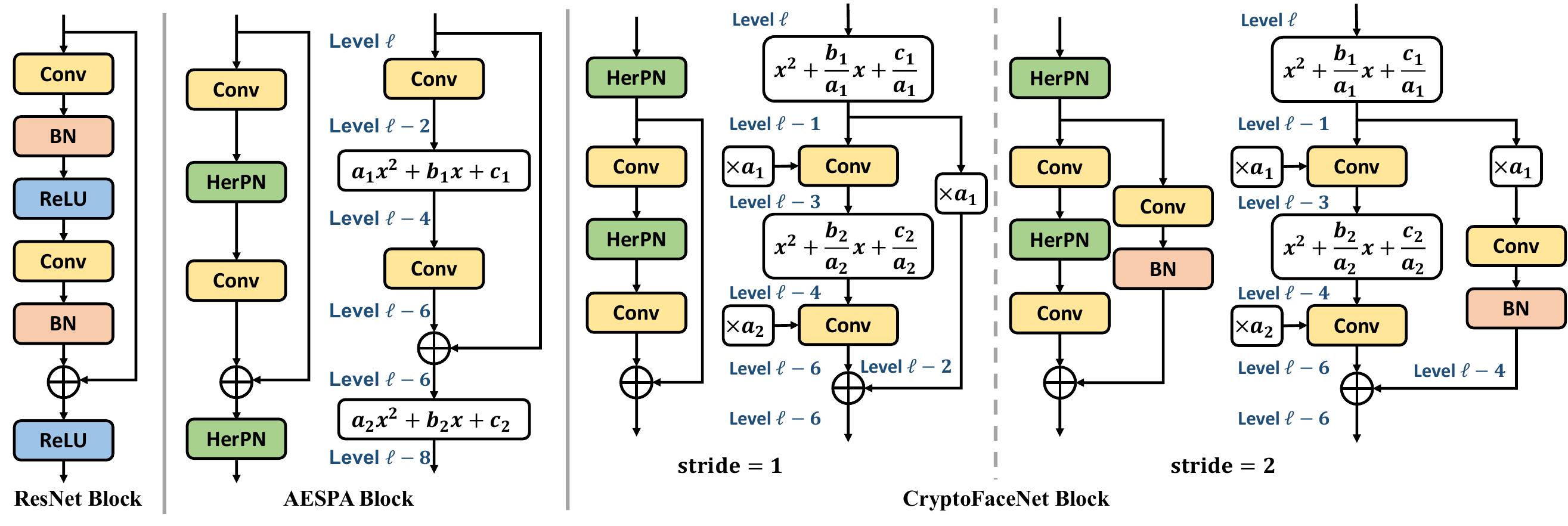}
    \vspace{-0.3cm}
    \caption{Convolutional blocks and their FHE implementations. Left: ResNet~\cite{he2016deep}; middle: AESPA~\cite{park2022aespa}; right: \cryptofacenet{}. \label{fig:block}} 
    \vspace{-0.4cm}
\end{figure*}

\subsection{Homomorphic Architecture \label{sec:arctecture}}
\textbf{Convolutional Block.}  As discussed in~\Cref{sec:hcnns}, we adopt FHE convolution from MPCNN~\cite{lee2022low} and the Hermite polynomial activation HerPN introduced by AESPA~\cite{park2022aespa}. \Figref{fig:block} shows AESPA block and its FHE implementation provided by~\cite{ao2023autofhe}. The AESPA block depth is 8 since one Conv consumes 2 levels, and one HerPN consumes 2 levels. We propose a depth-optimal \emph{shifted} AESPA block. \Figref{fig:block} shows \cryptofacenet{} blocks for $\operatorname{stride}=1$ and $\operatorname{stride}=2$. HerPN can be formulated as a degree-2 polynomial $ax^2+bx+c$ with depth 2. We fuse the coefficient $a$ to Conv weight and change the polynomial to $x^2+\frac{b}{a}x+\frac{c}{a}$ with depth 1. Therefore, the proposed \cryptofacenet{} block can save two levels. 

\noindent\textbf{Polynomial $\ell_2$ Normalization.} When computing the similarity score between two features $y_1$ and $y_2$, normalization of these features as $\frac{y_1}{\|y_1\|}$ and $\frac{y_2}{\|y_2\|}$ is necessary (\Eqref{eq:match}). $\ell_2$ normalization can be expressed as $\frac{y}{\| y \|_2}=y\cdot \frac{1}{\sqrt{\sum_{i=1}^d y_i^2}}$. Under FHE, $[y_i]^2$ can be computed via ciphertext-ciphertext multiplication, and the summation $\sum_{i=1}^d [y_i]^2$ can be obtained using rotations. However, the primary challenge lies in approximating the non-linear function $q(t)=\frac{1}{\sqrt{t}}$. This is difficult as the domain of $q(t)$ is typically very wide (since $t=|y|_2^2$). A Taylor expansion around $t = t_0$ would result in a polynomial with excessively high degree, and minimax approximation~\cite{lee2021minimax} is not directly applicable since $q(t)$ cannot be effectively scaled into the $[-1,1]$ interval it requires. Observing that $q(t)=\frac{1}{\sqrt{t}}$ is a strictly decreasing function and we propose using a simpler polynomial approximation: $p(t)=\beta_2t^2+\beta_1 t+ \beta_0$ to approximate $q(t)$. Unlike the minimax approximation that explicitly minimizes $|p(t)-q(t)|$, we instead ensure $p(t)$ has a similar shape to $q(t)$ within its relevant domain. We achieve this by selecting three control points $t_1$, $t_2$, and $t_3$, along with their corresponding reference values $\frac{1}{\sqrt{t_1}}$, $\frac{1}{\sqrt{t_2}}$, and $\frac{1}{\sqrt{t_3}}$, respectively. Solving these equations yields coefficients $\beta_2$, $\beta_1$, and $\beta_0$ that fit our polynomial to the distribution of data. Specifically, we set the control points based on the distribution of $t$ as follows: $t_1=\mathrm{Mean}(t)-\mathrm{Std}(t)$,  $t_2=\mathrm{Mean}(t)$, $t_3=\mathrm{Mean}(t)+\mathrm{Std}(t)$. Thus, the resulting polynomial $p(t)$ is distribution-aware, and its coefficients ($\beta_2$, $\beta_1$, and $\beta_0$) are determined by the underlying data distribution—similar to the score thresholding procedure. The polynomial approximation has a multiplicative depth of only 2, ensuring computational efficiency. Consequently, we estimate the matching threshold (i.e., $T$ in~\Eqref{eq:match}) to accommodate the polynomial-based $\ell_2$ normalization. 

\noindent \textbf{Adaptive Average Pooling.} MPCNN~\cite{lee2022low} originally employs adaptive average pooling with an output size of $(1, 1)$. To better preserve structural information crucial for FR, we customized it to have an output size of $(2, 2)$, resulting in $256$-dimensional features. Our $(2, 2)$ pooling is implemented as four separate $(1, 1)$ pooling operations, which permute elements of the feature vector. So, we also rearranged the fusion matrix $A$ accordingly (\Eqref{eq:fusion}) in the cleartext domain. 

\begin{figure}[t]
    \centering
    \includegraphics[width=\linewidth]{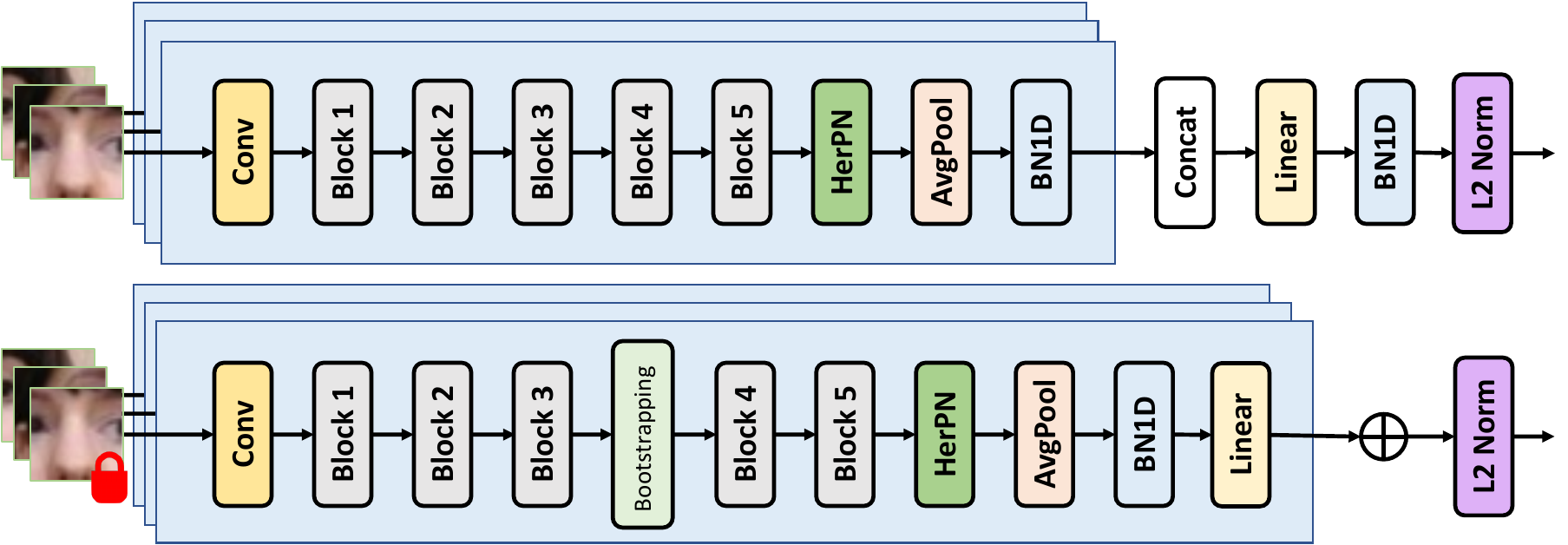}
    \caption{\cryptofacenet{}. Top: training; bottom: inference.  \label{fig:network}}
    \vspace{-0.5cm}
\end{figure}

\noindent\textbf{\cryptofacenet{} Architecture} is shown in \Figref{fig:network}. To reduce depth consumption, \cryptofacenet{} fuses the Linear and BatchNorm1D layers to a single Linear. The aggregation of features is a simple ciphertext addition. \cryptofacenet{} only uses one bootstrapping operation.
\section{Experiments \label{sec:experiments}}

\begin{table*}[tb]
\begin{center}
\resizebox{0.8\linewidth}{!}{
\begin{tabular}{l?llcc?cccccc?c?c}
    \topline
    \multirow{2}{*}{\textbf{Method}} & \multicolumn{4}{c?}{\textbf{Backbone}} & \multicolumn{6}{c?}{\textbf{Dataset}} & \multirow{2}{*}{\small \textbf{Latency}(s)} & \multirow{2}{*}{\small{\textbf{RAM}}}  \\ 
    
    & Network & \small{Params} &  \small{\#Boot} & \footnotesize{Res} & LFW & \small{AgeDB} & \small{CALFW} & \small{CPLFW} & \small{CFP-FP}  & Avg &  &  \\

    \midtopline 
    \multirow{2}{*}{MPCNN~\cite{lee2022low}} & \small{ResNet32} & 0.53M & 31 & 64 & 97.02 & 83.02 & 87.00 & 78.90 & 82.07 & 85.60 &  7,367 & 286G \\

     & \small{ResNet44} & 0.72M & 43 & 64 & 98.27 & 87.45 & 90.85 & 83.72 & 87.90 & 89.64 &  9,845 & 286G \\ \midtopline 

    AutoFHE~\cite{ao2023autofhe} & \small{ResNet32} & 0.53M & 8 & 64 & 93.53 & 80.88 & 85.40 & 75.67 & 77.96 & 82.69 &  4,001 & 286G  \\ \midtopline

    \multirow{3}{*}{\textbf{\ourface{}}}  & \small{\cryptofacenet{}4} & 0.94M & \multirow{3}{*}{1} & 64 & 98.87 & 89.45 & 91.60 & 81.98 & 85.21 & 89.42 &  1,364 & 269G \\

    & \small{\cryptofacenet{}9} & 2.12M & & 96 & 99.18  & 91.38 & 93.32 & 84.23 & 86.81 & 90.99 & 1,395  & 276G \\

    & \small{\cryptofacenet{}16} & 3.78M & & 128 & 98.78  & 92.90  &  93.73 & 83.95 & 87.94 & 91.46 &   1,446 & 277G \\
    
    \bottomline
\end{tabular}
}
\end{center}
\vspace{-0.5cm}
\caption{Experiments on end-to-end FR on FHE. \label{tab:result}}
\vspace{-0.2cm}
\end{table*}

\begin{table*}[tb]
\begin{center}
\resizebox{\linewidth}{!}{
\begin{tabular}{l?l?cccccccccc}
    \topline

    \textbf{Method} & \small{Backbone} & Conv & BN & Residual & AvgPool & Linear & Activation & L2 Norm & Match & \small{Bootstrapping} & Other   \\

    \midtopline 

    \multirow{2}{*}{MPCNN~\cite{lee2022low}} & \small{ResNet32} & 896s (12.17\%) & 28s (0.38\%) & 0.5s (0.01\%) & 46s (0.62\%) & 807s (10.96\%) & 583s (7.92\%) & 5s (0.07\%)  & 2s (0.02\%) & 4991s (67.76\%) & 7s (0.09\%)  \\

     & \small{ResNet44} & 1214s (12.33\%) & 39s (0.39\%) & 0.7s (0.01\%) & 46s (0.46\%) & 807s (8.19\%) & 807s (8.20\%) & 5s (0.05\%) & 2s (0.02\%) & 6917s (70.27\%) & 7s (0.08\%)  \\ \midtopline 

    AutoFHE~\cite{ao2023autofhe} & \small{ResNet32} & 1966s (49.14\%) & 28s (0.69\%) & 1.7s (0.04\%)  &  38s (0.95\%) & 658s (16.43\%) & 17s (0.43\%)  & 4s (0.10\%) & 1s (0.03\%) & 1274s (31.84\%)  & 14s (0.34\%)   \\ \midtopline

     \textbf{\ourface{}}  & \footnotesize{\cryptofacenet{}4} & 858s (62.93\%) & 2s (0.16\%) & 0.1s (0.01\%) & 26s (1.94\%) & 277s (20.30\%) & 13s (0.94\%) & 2s (0.11\%) & 0.3s (0.02\%) & 141s (10.34\%) & 44s (3.26\%)  \\
    \bottomline
\end{tabular}
}
\end{center}
\vspace{-0.5cm}
\caption{Latency of operations on FHE for $64\times64$ encrypted face images. \label{tab:latency}}
\vspace{-0.5cm}
\end{table*}

\textbf{Datasets.} (1) Training dataset: we use WebFace4M, a subset of WebFace260M~\cite{zhu2021webface260m} to train \ourface{} and baselines on cleartext data in Pytorch. WebFace4M has 205,990 identities and 4,235,242 face images in total. (2) Test datasets: we follow AdaFace~\cite{kim2022adaface} to benchmark \ourface{} and baselines on five standard test datasets~\cite{insightface}: LFW~\cite{huang2008lfw}, AgeDB~\cite{moschoglou2017agedb}, CALFW~\cite{zheng2017calfw}, CPLFW~\cite{zheng2018cplfw}, CFP-FP~\cite{sengupta2016cfpfp}. The image resolution is $112\times112$. In our experiments, we resize face images to small ($64\times 64$), medium ($96 \times 96$), and high-resolution ($128 \times 128$), which satisfies use cases corresponding to edge FR, embedded FR, and cloud FR in the wild. CFP-FP has 7,000 pairs, resulting in 14,000 face images, while others have 6,000 pairs, resulting in 12,000 face images. We report the one-to-one verification accuracy on the encrypted test datasets. 

\noindent\textbf{FHE Library and Hardware.} We follow MPCNN~\cite{lee2022low} and AutoFHE~\cite{ao2023autofhe} to adopt SEAL~\cite{seal3.6} library and report latency and RAM footprint on Amazon AWS r5.24xlarge. The modified SEAL~\cite{lee2022low} incorporates bootstrapping.

\noindent\textbf{Baselines.} As there is no prior end-to-end encrypted FR, we adopt two FHENets MPCNN and AutoFHE (see~\Cref{sec:hcnns}) as our baselines because they report the SoTA performance on CIFAR image classification. Our C++ implementation of \ourface{} is built on top of MPCNN and AutoFHE. To take $64\times 64$ face images as input, we change the stride of the very first convolutional layer from 1 to 2 to ensure that a single ciphertext can pack any intermediate tensors. The new output layers used for FR are AdaptiveAvgPool2d((2, 2)) $\mapsto$ Linear(256, 256) $\mapsto$ BatchNorm1d(256).  AutoFHE is a search-based approach and reports search results on the CIFAR dataset. Extending the search algorithm from a small CIFAR dataset to the much larger WebFace dataset is challenging, so we use the search result on CIFAR and transfer it to the WebFace dataset.

\noindent\textbf{Parameters.} (1) Training: we adopt the parameters from AdaFace~\cite{kim2022adaface} without tuning. We set the learning rate to 0.05, epochs to 26, batch size to 256, momentum to 0.9, and weight decay to 0.0005. We use the SGD optimizer with multi-step scheduler. The learning rate is scaled by $0.1\times$ at epochs 12, 20, and 24. When we train \ourface{} and AutoFHE, we clip the gradient to 1 as suggested by AutoFHE. The patch size is set to $32\times 32$. We set the strength ($\alpha$) of the jigsaw loss to  0.005. (2) FHE: To meet 128-bit security~\cite{cheon2019hybrid}, we use the same CKKS parameters as MPCNN and AutoFHE. The degree of the cyclotomic ring is $2^{16}$ and Hamming weight is 192. We set the default modulus to 46 bits and the special modulus to 51 bits~\cite{lee2022low,ao2023autofhe}. 

\subsection{End-to-End Encrypted FR on FHE}

We follow the standard 10-fold cross-validation~\cite{deng2022arcface,kim2022adaface} to benchmark \ourface{} and baselines on the encrypted face datasets. For each split, nine groups (cleartext) are used to estimate the threshold and polynomial approximation of $\ell_2$ normalization, while the standalone group (ciphertext) is used to test performance. We report the average verification accuracy of 10-fold cross-validation for each dataset as shown in Table~\ref{tab:result}. For a pair of face images, we consider the first as the reference and the second as the probe. Each experiment includes an offline and an online stage (see \Cref{sec:cryptoface}). In Table~\ref{tab:result}, we report the online latency. We use numbers of patches (\ie 4, 9, and 16) to denote \cryptofacenet{} for different resolutions $64\times 64$, $96\times 96$, and $128 \times 128$, respectively. 

For encrypted face images at resolution $64\times 64$,  MPCNN with ResNet44 shows the highest accuracy (89.64\%) but a prohibitively large latency (9,845 seconds). \ourface{} greatly accelerates encrypted face recognition by $\bm{7.2\times}$ which translates to savings of 8,481 seconds. We only observe a negligible accuracy drop of $0.22\%$. MPCNN with ResNet32 is a faster version but only achieves $85.60\%$ accuracy. \ourface{} speeds up inference by $\bm{5.4\times}$ and increases FR performance by $\bm{+3.82\%}$. AutoFHE is much faster than MPCNN. However, the transferred AutoFHE achieves $82.64\%$ on encrypted FR with a latency of 4,001 seconds. Compared to AutoFHE, \ourface{} accelerates inference by $\bm{2.9\times}$ and improves the encrypted FR performance by $\bm{+6.73\%}$. \ourface{} also reduces RAM footprint by \textbf{17G} since we only need one bootstrapping operation, while the baselines require three operations for different repeated packing copies $M$ (see \Cref{sec:hcnns}). The experimental results demonstrate the effectiveness of \ourface{}, the proposed end-to-end encrypted FR. The proposed \cryptofacenet{} is an efficient FHENet thanks to the mixture of PCNNs with simple, yet effective, parallelization.  

We also analyze how the resolution of encrypted face images impacts verification accuracy, latency, and RAM footprint, as shown in Table~\ref{tab:result}. When the resolution increases from $64\times64$ to $96\times96$ and $128\times 128$, \ourface{} can take advantage of high-resolution face images to effectively improve FR performance. Compared to \cryptofacenet{}4, \cryptofacenet{}9  and \cryptofacenet{}16 increase the accuracy by $\bm{+1.57\%}$ and $\bm{+2.04}\%$, respectively. \ourface{} can maintain a nearly constant latency (1,364 to 1,446 seconds) even as the image resolution increases from $64\times 64$ to $128 \times 128$. \ourface{} slightly increases the RAM footprint from 269G to 277G. The results demonstrate that \ourface{} is scalable to high-resolution images and can satisfy different requirements in real-world secure FR applications.

\subsection{Operation Latency}

Table~\ref{tab:latency} lists detailed latency of different operations. Bootstrapping operations dominate the latency of MPCNN, around $70\%$. AutoFHE successfully removes most bootstrapping operations by using mixed-degree polynomial activations. \ourface{} fundamentally decreases the depth of networks and significantly reduces the number of bootstrapping operations to \emph{one} with bootstrapping only contributing $10\%$ to the total latency. The parallelism of \cryptofacenet{} on FHE introduces an acceptable overhead, around $<3.26\%$ (Other). \ourface{} and MPCNN (ResNet32) spend roughly equal time evaluating convolutional layers. However, MPCNN (ResNet32) has 31 convolutional layers, and one patch CNN only has 13 (including two residual convolutional layers) since \cryptofacenet{} convolutional layers take high-level ciphertext as input, leading to higher latency. Both AutoFHE and \ourface{} benefit from low-degree polynomial activations because expensive ciphertext-ciphertext multiplications are decreased. We apply the average pooling with the output size $(2, 2)$ to get 256-dimensional features (see~\Cref{sec:arctecture}). The $(2, 2)$ average pooling is slower than $(1, 1)$ averaging pooling used by the original versions of MPCNN and AutoFHE because it introduces more rotations and multiplications. However, FR necessitates high-dimensional features to preserve more structural information. The proposed polynomial approximation of $\ell_2$ normalization and the feature matching are very efficient under FHE. \cryptofacenet{} is a depth-optimized homomorphic neural architecture (see~\Cref{sec:arctecture}) demonstrating lower latency over different FHE operations.

\subsection{Polynomial L2 Approximation \label{sec:polynomial}}

\begin{figure}[t]
    \centering
    \includegraphics[width=\linewidth]{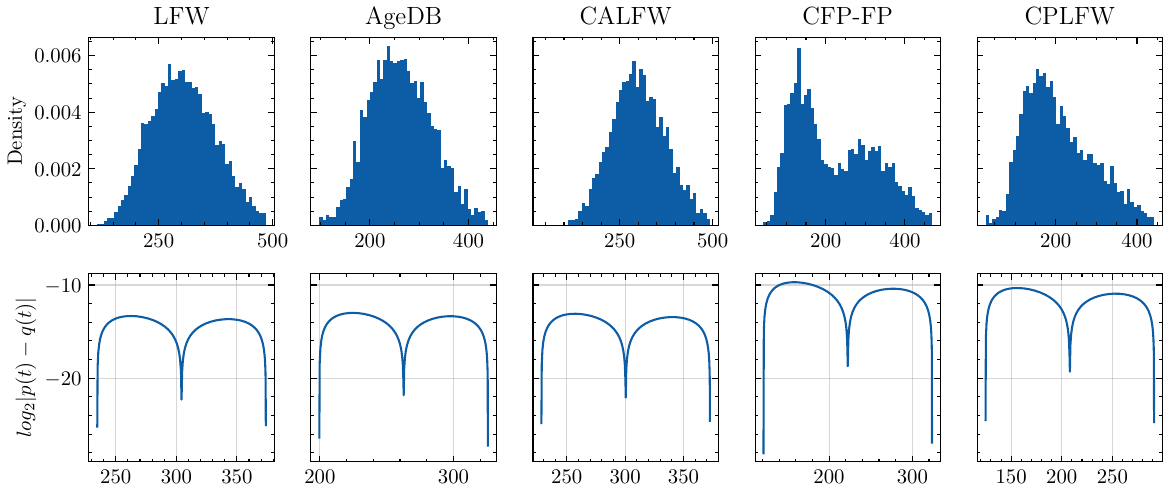}
    \caption{$\ell_2$ polynomial approximation. Top: the distribution (domain) of $q(t)=1/\sqrt{t}$, bottom: the approximation error $\log_2{|p(t)-q(t)|}$. \label{fig:polynomial}}
    \vspace{-0.5cm}
\end{figure}

\Figref{fig:polynomial} (top) shows the distribution of $\|y\|_2^2$, namely the domain of $q(t)=1/\sqrt{t}$. We propose a distribution-aware polynomial approximation $p(t)\mapsto q(t)$ (see~\Cref{sec:arctecture}). \Figref{fig:polynomial} (bottom) shows that the approximation error is $\log_2|p(t)-q(t)|\leq 2^{-10}$. Table~\ref{tab:latency} shows the latency of the polynomial $\ell_2$ approximation is $0.3$ to $2$ seconds on FHE only consuming $0.02\%$ of inference time. Thus, the polynomial $\ell_2$ approximation is accurate and efficient.

\subsection{Mixed-Quality FR Benchmarks}

\begin{figure}[t]
    \centering
    \begin{minipage}{0.24\textwidth} 
        \centering
        \includegraphics[width=\textwidth]{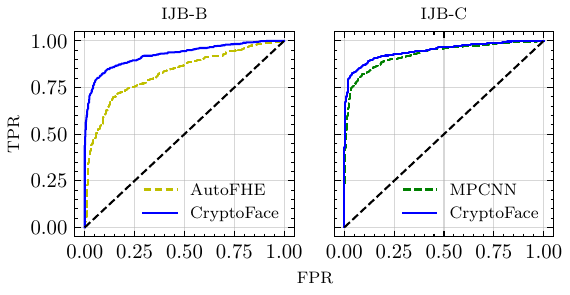}
    \end{minipage}%
    \hspace{0.1pt} 
    \begin{minipage}{0.22\textwidth}
        \centering
        \resizebox{\textwidth}{!}{
        \begin{tabular}{c|c|l} 
            \toprule
            IJB-B & AutoFHE & \textbf{\ourface{}} \\ 
            Time & 4,001s & 1,364s (\textcolor{teal}{$\bm{\downarrow 2,637}$s}) \\ 
            AUC & 0.825 & 0.926 (\textcolor{teal}{$\bm{\uparrow 0.1}$}) \\ 
            \midrule
            IJB-C & MPCNN & \textbf{\ourface{}} \\ 
            Time & 7,367s & 1,364s (\textcolor{teal}{$\bm{\downarrow 6,003}$s}) \\
            AUC & 0.927 & 0.946 (\textcolor{teal}{$\bm{\uparrow 0.02}$}) \\
            \bottomrule
        \end{tabular}
        }
    \end{minipage}
    \vspace{-0.3cm}
    \caption{Experiments on IJB-B and IJB-C. \label{fig:ijb}}
    \vspace{-0.5cm}
\end{figure}

The five standard FR datasets used in our experiments are regarded as high-quality FR benchmarks. They can satisfy the most real-world secure FR applications. IJB-B~\cite{ijbb} and IJB-C~\cite{ijbc} FR datasets include mixed-quality face images and are used to test FR models for challenging FR tasks. \Figref{fig:ijb} shows experimental results on encrypted IJB-B and IJB-C on FHE. Due to the prohibitively high computational cost, we randomly sample 1,200 pairs (with 50\% positive pairs and 50\% negative pairs) from each dataset. We use ROC (Receiver Operating Characteristic) curve and AUC (Area Under the Curve) to qualify the FR performance of \ourface{} and baselines. From~\Figref{fig:ijb}, \ourface{} consistently shows better performance on IJB-B or IJB-C on FHE compared to AutoFHE and MPCNN.  

\subsection{Identification}
\begin{wrapfigure}{r}{0.55\linewidth}
    \centering
    \vspace{-0.45cm}
    \resizebox{\linewidth}{!}{
    \begin{tabular}{l|cc}
        \toprule
        Method & Rank-1 Acc. & Rank-5 Acc.  \\
        \midrule
        MPCNN & 88.28 & 97.66  \\
        \textbf{\ourface{}} & 92.19 (\textcolor{teal}{$\bm{\uparrow 3.91\%}$}) & 98.44 (\textcolor{teal}{$\bm{\uparrow 0.78\%}$})  \\
        \bottomrule
    \end{tabular}}
    \vspace{-0.2cm}
    \captionof{table}{1:128 closed-set retrieval. \label{tab:retrieval}}
    \vspace{-0.3cm}
\end{wrapfigure}
In this paper, we primarily focus on the face verification task, as discussed in~\Cref{sec:cryptoface}. However, \ourface{} can be readily extended to face identification scenarios, such as one-to-many face matching. In Table~\ref{tab:retrieval}, we report experimental results for a $1:128$ closed-set rank retrieval task under FHE, using a randomly selected subset of LFW consisting of 128 pairs. The latency difference solely arises from the feature extraction stage, as previously reported. We evaluate performance using Rank-1 and Rank-5 accuracy as metrics. The experimental results confirm that \ourface{} is also effective for face identification tasks.
\section{Conclusion}

This paper introduced \ourface{}, the first end-to-end encrypted face recognition system using fully homomorphic encryption (FHE). Once face images are encrypted, all subsequent operations like feature extraction, matching, and score thresholding are performed in the encrypted domain without decryption. The key idea behind \ourface{} is \cryptofacenet{}, a novel architecture which is a mixture of shallow patch convolutional neural networks optimized for FHE compatibility and mitigating the steep computational burden of encrypted inference. Experimental results on standard face recognition benchmarks show that \ourface{} is $7\times$ faster than SoTA FHENets while achieving better verification performance. \ourface{} can effectively process high-resolution encrypted face images to improve verification accuracy by $+2\%$ while maintaining near-resolution-independent latency. \ourface{} will facilitate the deployment of secure face recognition systems in applications requiring strict privacy and security guarantees.

{
    \small
    \bibliographystyle{ieeenat_fullname}
    \bibliography{egbib_nolink}
}

\end{document}